\documentclass[final]{cvpr}

\usepackage{times}
\usepackage{epsfig}
\usepackage{graphicx}
\usepackage{amsmath}
\usepackage{amssymb}

\usepackage{tabularx}
\usepackage{color}
\usepackage{subfig}
\usepackage{booktabs}
\usepackage{amsthm}
\usepackage[T1]{fontenc}

\usepackage[pagebackref=true,breaklinks=true,colorlinks,bookmarks=false]{hyperref}

\def\cvprPaperID{13} 
\def\confYear{CVPR 2021}

\begin{document}

\title{Image-Based Plant Wilting Estimation}



\author{\parbox{16cm}{\centering
    {Changye Yang$^{\star}$ \quad Sriram Baireddy$^{\star}$ \quad Enyu Cai$^{\star}$ \\ Val\'erian M\'eline$^{\dagger}$ \quad Denise Caldwell$^{\dagger}$ \quad Anjali S. Iyer-Pascuzzi$^{\dagger}$  \quad Edward J. Delp$^{\star}$}\\
    {\normalsize
    $^{\star}$ Video and Image Processing Lab (VIPER), Purdue University, West Lafayette, Indiana, USA\\
    $^{\dagger}$ Iyer-Pascuzzi Lab, Purdue University, West Lafayette, Indiana, USA
    }
}}

\maketitle

\begin{abstract}
	Many plants become limp or droop through heat, loss of water, or disease. This is also known as wilting.
	In this paper we examine plant wilting caused by bacterial infection.
	In particular we want to  design a metric for wilting based on images acquired of  the plant.
    A quantifiable wilting  metric  will be useful in studying bacterial wilt and identifying resistance genes. 
    Since there is no standard way to estimate wilting, it is common to use ad hoc visual scores. 
	This is very subjective and requires expert knowledge of the plants and the disease mechanism.  
	Our solution consists of using various wilting metrics acquired from RGB images of the plants.
	We also designed several experiments to demonstrate that our metrics are effective at estimating wilting in plants.
\end{abstract}

\begin{figure}[t]
\centering

\subfloat[]{\label{Figure:input1}{\includegraphics[width = 0.4\textwidth]{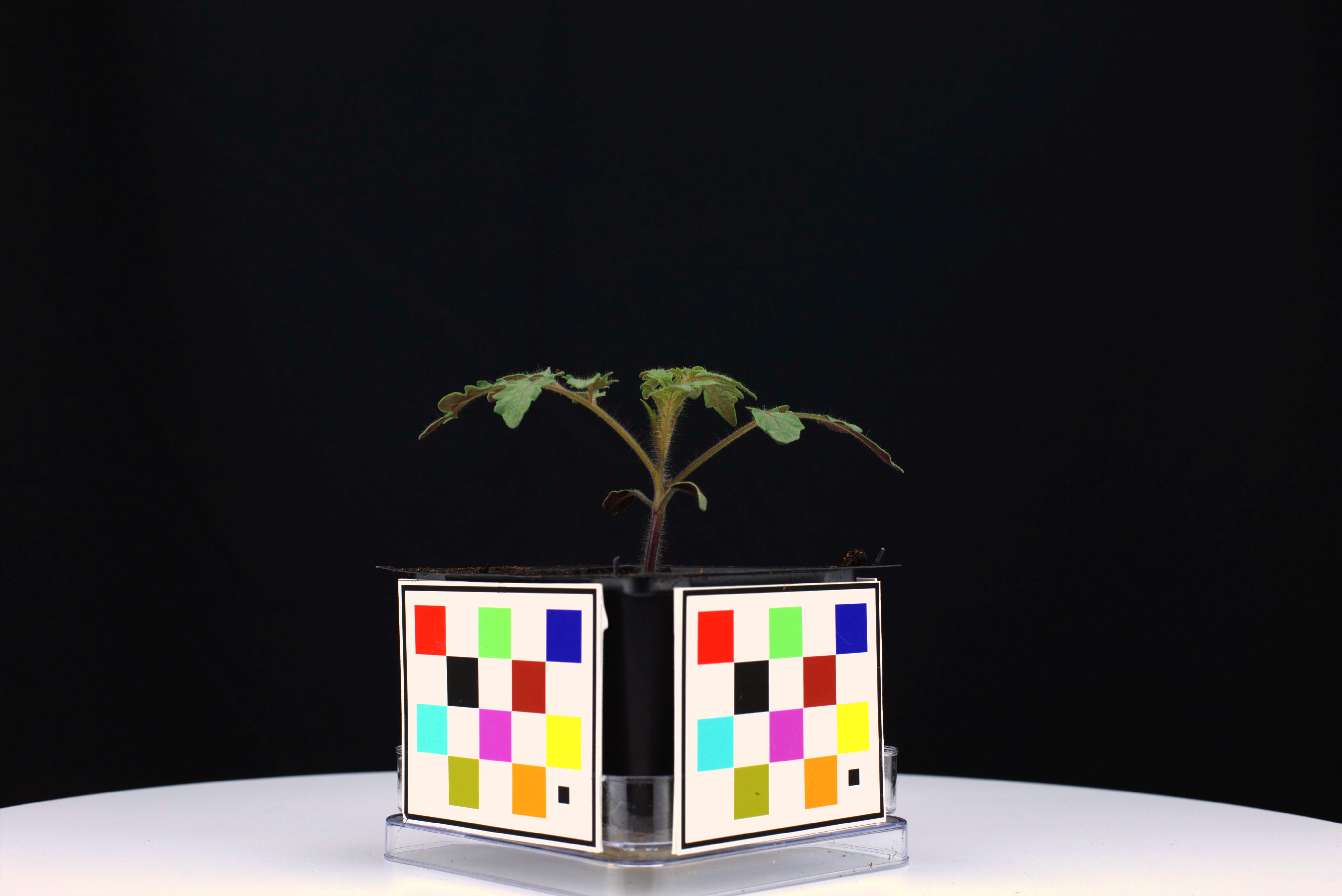}}}
\quad
\subfloat[]{\label{Figure:input1}{\includegraphics[width = 0.4\textwidth]{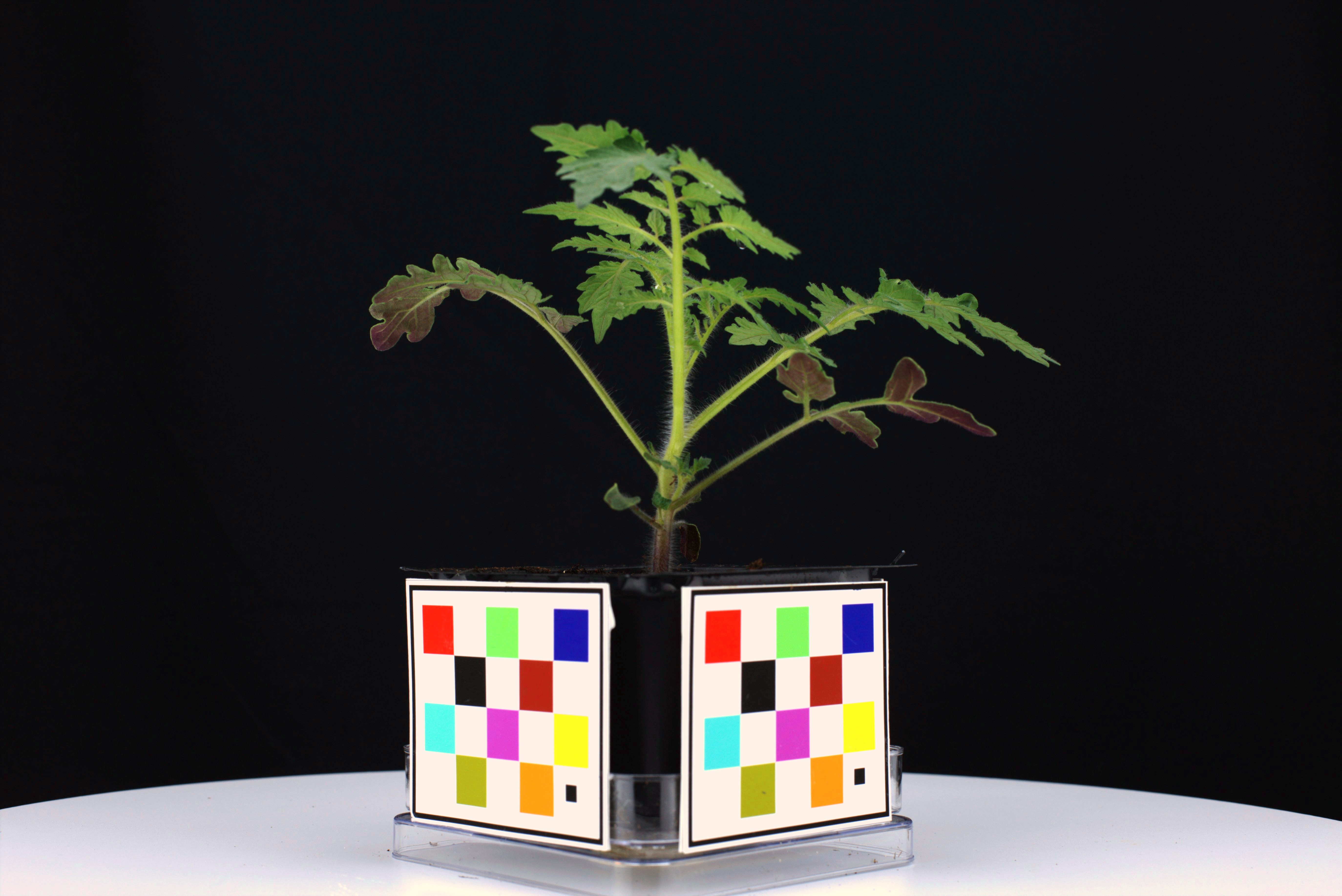}}}

    
    \caption{An example of the growth of a plant without bacterial inoculation
             (a) -1 dpi (\emph{days post inoculation})
             (b) 3 dpi
             }
    \label{orig_healthy}
\end{figure}

\begin{figure*}[]
	\centering
	\centerline{\includegraphics[width = 1.0\textwidth]{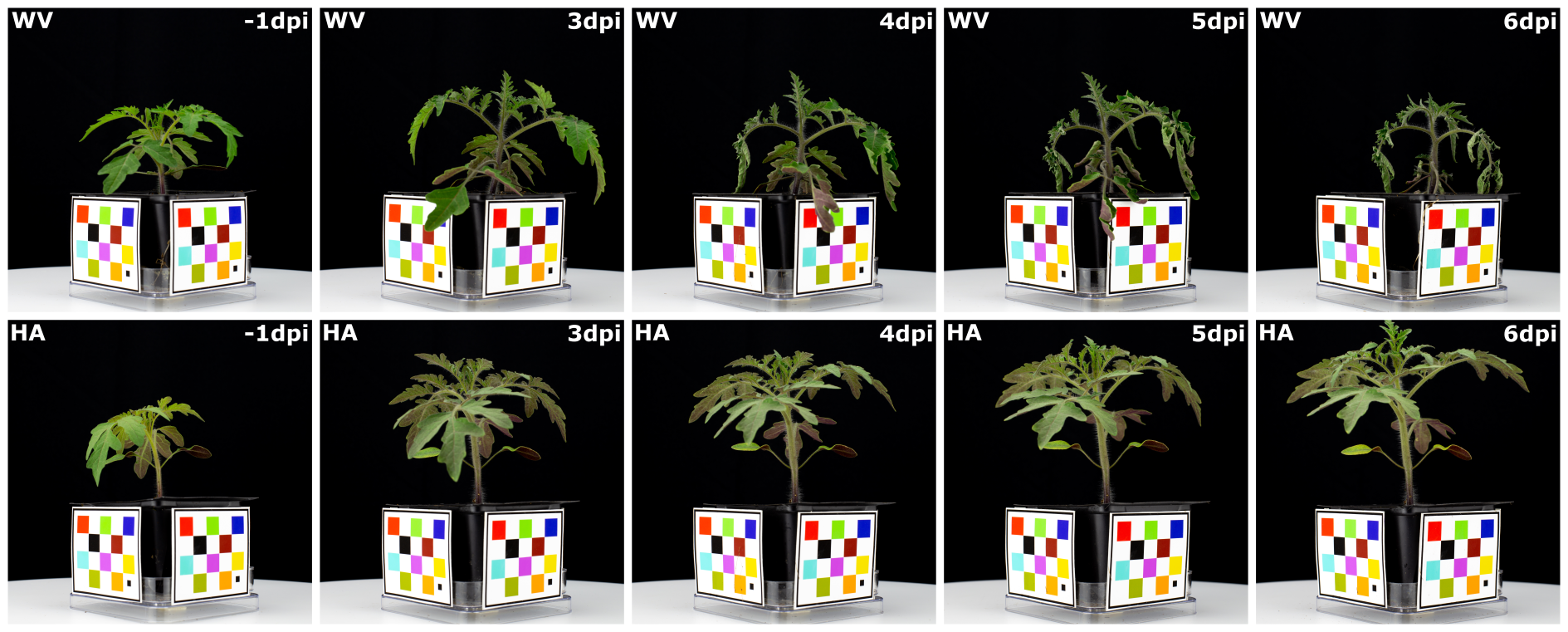}}
	\caption{Plants inoculated with \textit{Rs} bacteria. Here, \emph{dpi} means \emph{days post inoculation}. The Hawaii 7996 (HA) plants are resistant to infection while the West Virginia 700 (WV) plants are susceptible to infection.}
\label{plt_evo}
\end{figure*}
\section{Introduction}



\textit{Ralstonia solanacearum} (\textit{Rs})~\cite{Agrios1997} is a soil-borne bacterium that infects plant roots and ultimately causes susceptible plants to wilt and die.
Bacterial wilt caused by \textit{Rs}~\cite{Agrios1997} is a major threat to crop production around the world~\cite{guji2019,ji2005,zinnat2018}. 
In the United States, some of the most important vegetable crops, such as tomatoes and potatoes~\cite{USDA2012}, can  suffer significant yield loss from bacterial wilt~\cite{aslam2017,zinnat2018}. 
Figure~\ref{orig_healthy} shows an example of a healthy tomato plant, while 
Figure~\ref{plt_evo} shows the growth of tomato plants exhibiting different levels of resistance after inoculation with the bacterium \textit{Rs}.

One of the best bacterial wilt controlling methods is through developing genetically resistant plants, but resistance genes that act on US strains of \emph{Rs} have not been identified~\cite{huet2014}.
To identify resistance genes, metrics to quantify \emph{Rs}-induced plant wilt are desirable. 
In this paper, we propose a set of computational wilting metrics using several RGB images of the plant. 
We tested our metrics on high \textit{Rs} resistance and low \textit{Rs} resistance tomato plants, both with and without bacterial infection. 
In addition, we use machine learning-based methods to show that expert-labeled wilting scores can be predicted using our metrics.


\begin{figure}[]
	\centering
	\centerline{\includegraphics[width = 0.5\textwidth]{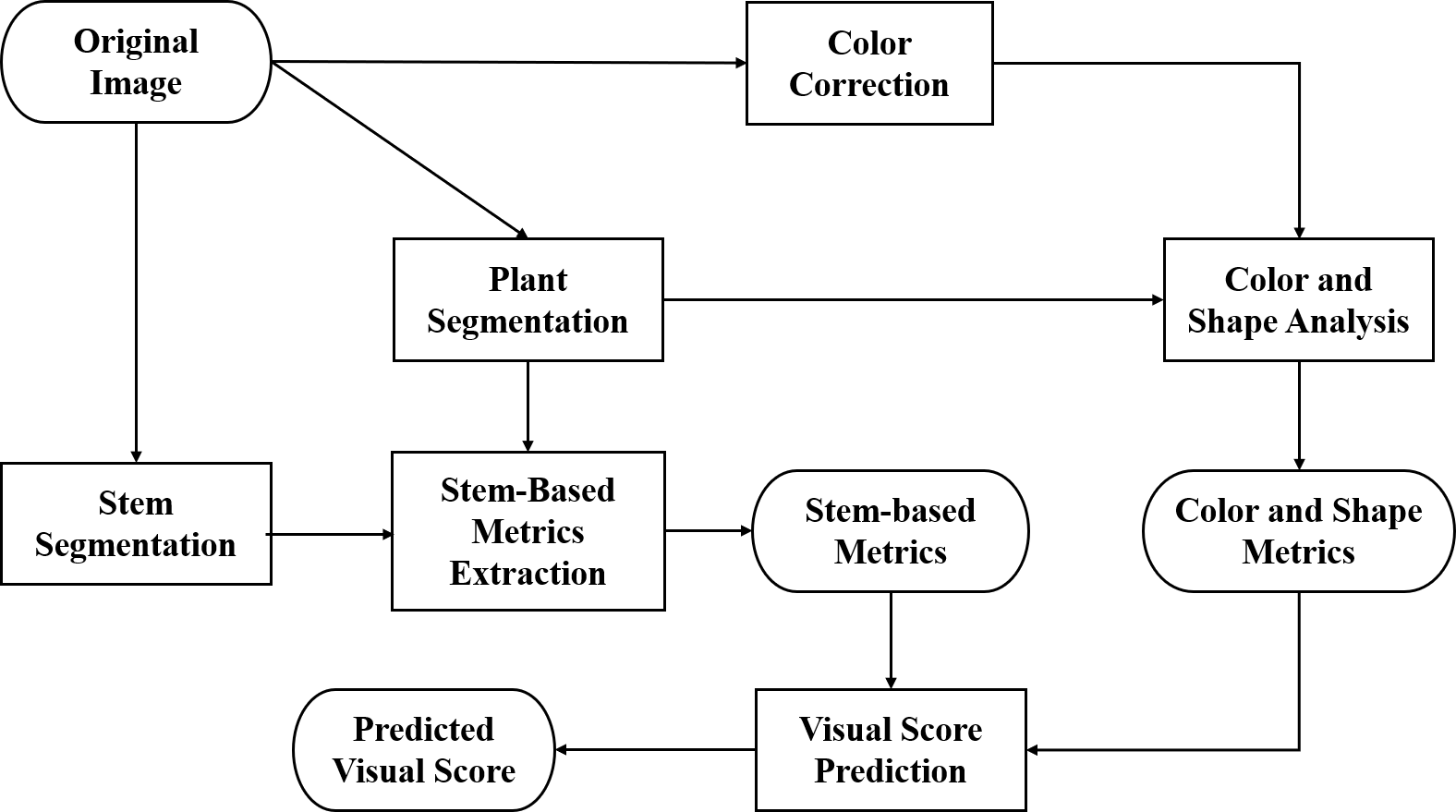}}
	\caption{Block Diagram for Our Method}
\label{Block_Diagram}
\end{figure}
\section{Review and Assumptions}
There are many ways for plant scientists to quantify the effect of plant diseases~\cite{brecht2007,ji2005,kover2002,lacroix2016}, but each method has its own shortcomings. 
Some studies use crop yield (amount of a crop grown per unit area of land) as the metric~\cite{fischer2015}.
In~\cite{ji2005}, Ji \etal use  crop yield as an indicator to study the effectiveness of using the chemical Thymol as a biofumigant to control \textit{Rs}-induced tomato bacterial wilt. 
In~\cite{kover2002}, Kover \etal use seed production to measure pathogen resistance on \textit{Arabidopsis thaliana}. 
The downside of using crop yield is that it requires long waiting periods for the plants to be ready for harvesting.
Many studies also measure the amount of pathogen directly from samples of the plants~\cite{fang2015,lacroix2016,malarczyk2019}. 
To do this, either the whole plant or parts of the plant are collected to be examined for the presence of the pathogen.
This destructive sampling method often hinders plant growth and sometimes even kills the plants. 
In addition, both crop yield and destructive sampling methods are difficult to use to show the long term wilting trend of the same plant. 
Another approach is to have experts visually examine the plants to determine wilting. 
For example, in \cite{brecht2007}, Engelbrecht \etal measure leaf water potential using visual assessment that is very subjective and difficult to reproduce. 
For tomato plants, experts rate each plant on its degree of wilting, taking into consideration many features of the plant such as the the overall loss of plant mass and the shift in color~\cite{aslam2017}.
In most cases only trained experts are capable of assigning a proper visual score.
We choose to design our wilting metrics using several RGB images of the plant. 
RGB images are commonly used for other plant studies~\cite{camargo2009,chen_2018,fan_2018,oerke2011}.  
Using RGB images is also non-destructive and easily accessible.

Several sensor-based wilting metrics in the past have been proposed.
In~\cite{caplan2019}, Caplan \etal use manually measured leaf angles as an indicator of drought stress. 
In~\cite{bock2020}, Bock \etal determine disease severity with RGB images of individual leaves. 
These methods require acquiring individual leaves so they are very labor intensive. 
Other methods such as~\cite{Mizuno2007,wakamori2019} also use RGB images for estimating wilting, but they require special equipment such as guided rail cameras and laser sensors~\cite{wakamori2019} or field servers~\cite{Mizuno2007}. 
Our method uses RGB images from multiple views using a single RGB camera which requires much less equipment and labor.  

In~\cite{sancho2019}, Sancho \etal use RGB image-based color information as part of the metrics to estimate \textit{Verticillium} wilt of olive plants. 
Sancho \etal incorporated eighteen color measurements into their metrics, but some of the metrics require cutting the olive leaves. 
Similar to Sancho \etal, the metric we proposed also uses RGB image-based color information, but we reduce the number of color based metrics from eighteen to one and our method does not require cutting physical leaves. 
We also add a color correction step to account for the difference in imaging conditions.

From our observations, bacterial wilt has significant impact on the color and shape of plants.
Examples can be seen in Figure~\ref{orig_healthy} and Figure~\ref{plt_evo}. 
Our primary approach is to estimate color and shape information from RGB images. 
For color information, we estimate the color distribution of the plant image pixels. 
For shape analysis, we first use the convex hull~\cite{barber1996} to capture the outline of the plant. 
Since the convex hull only captures information related to the outer edge of the plants, we developed several stem-based metrics which estimate the distribution of the plant materials relative to the stem.

\section{Proposed Wilting Metrics}
In this section, we first give an overview of our wilting metrics. 
For our experiment in Section~\ref{exper} we use two types of plants: (1) \textit{Solanum lycopersicum}, or Hawaii 7996 (HA), which is a high \textit{Rs} resistance plant; and (2) \textit{Solanum pimpinellifolium}, or West Virginia 700 (WV), which is a low \textit{Rs} resistance plant~\cite{french2018}. 
Both types of plants are divided into an ``inoculated'' (the bacteria \textit{Rs} is introduced to the plant via water) group and a ``mock'' (the control group, no \textit{Rs} is introduced) group.

\begin{figure}[]
    \centering
    \subfloat[]{\label{Figure:input1}{\includegraphics[width = 0.21\textwidth]{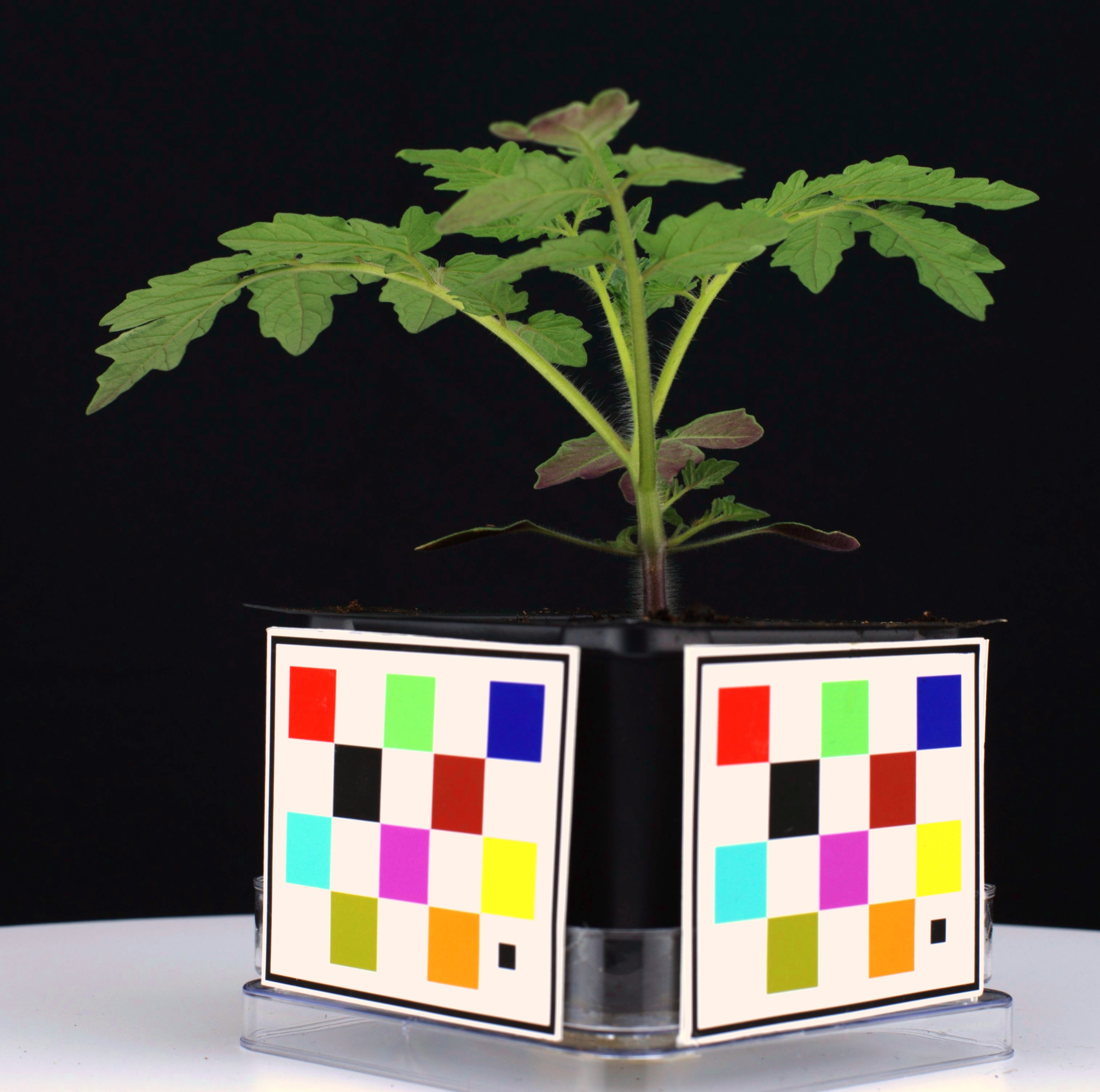}}}
    \quad
    \subfloat[]{\label{Figure:prob1}{\includegraphics[width = 0.21\textwidth]{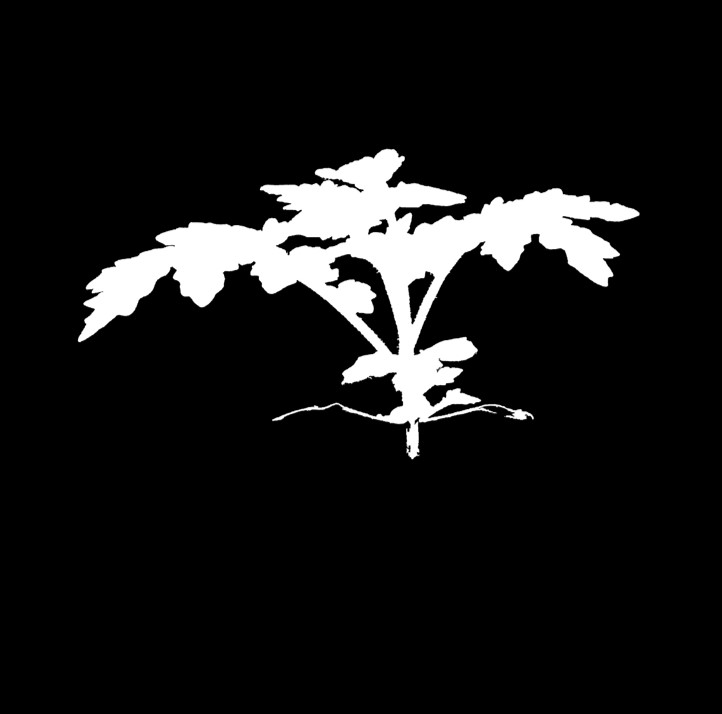}}}
    \qquad \qquad \qquad
    \subfloat[]{\label{Figure:input2}{\includegraphics[width = 0.21\textwidth]{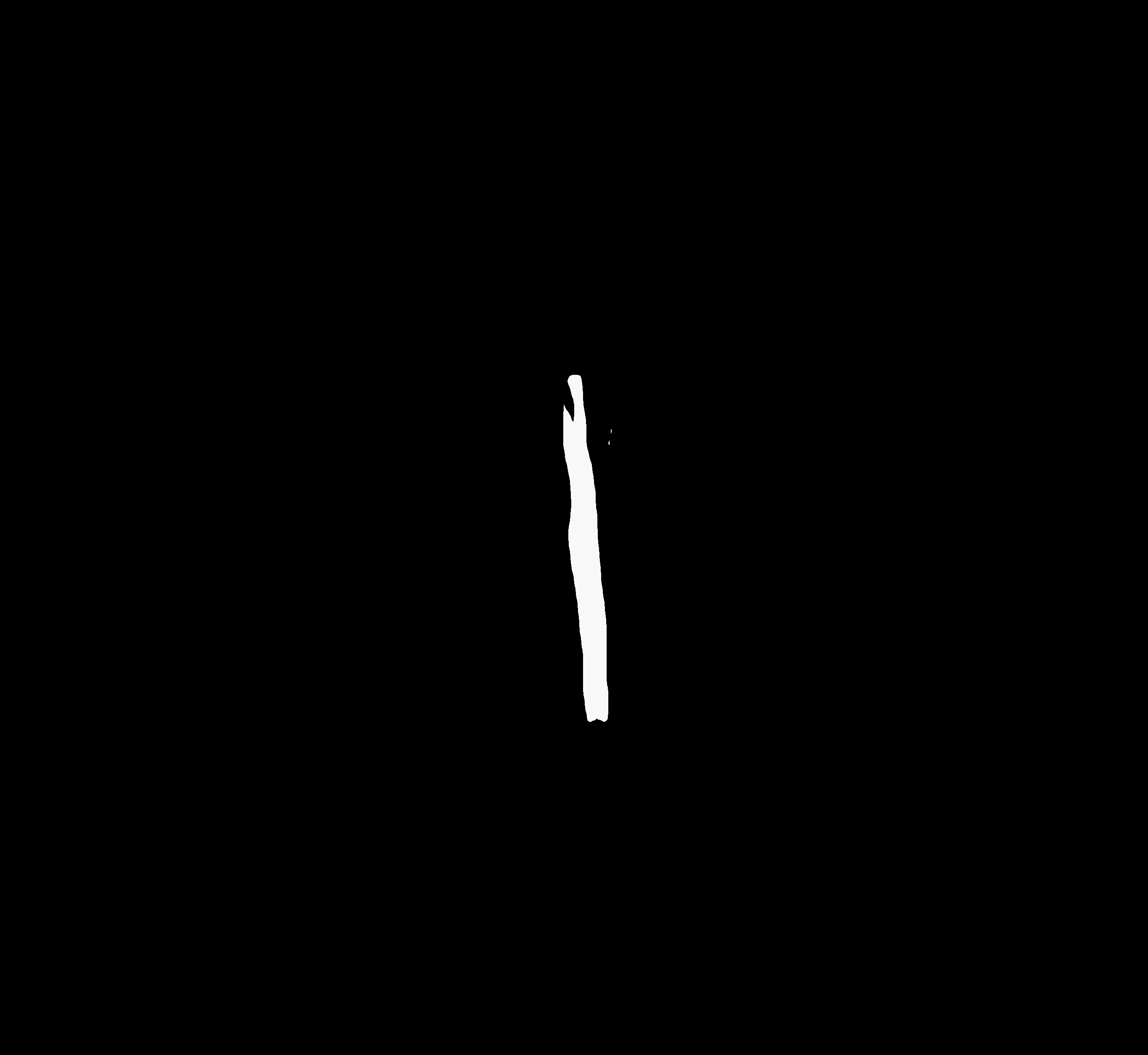}}}
    \quad
    \subfloat[]{\label{Figure:prob2}{\includegraphics[width = 0.21\textwidth]{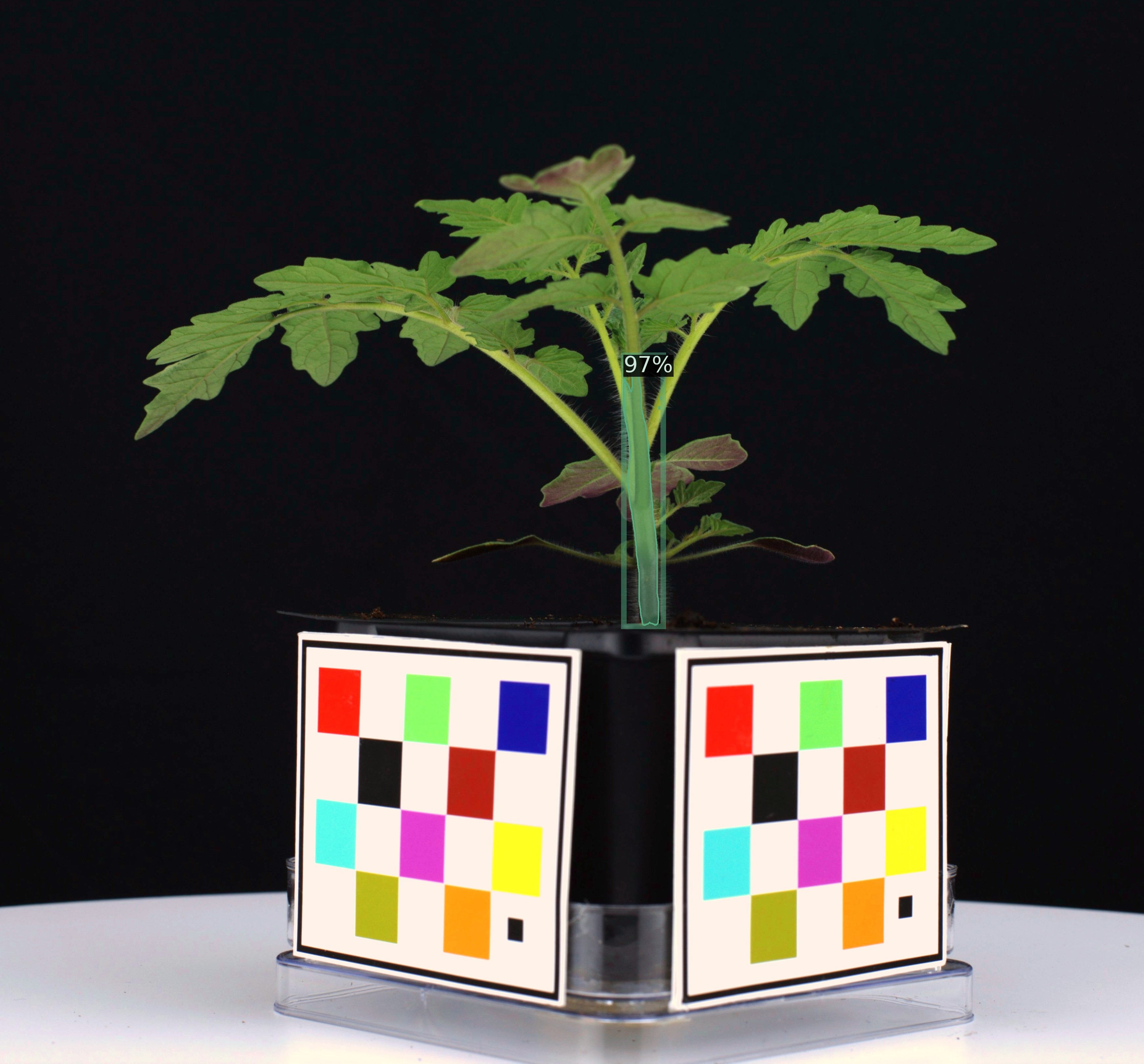}}}
    \qquad \qquad \qquad
    \caption{An example of a mock WV plant, all images are cropped : 
             (a) Original Plant (cropped).
             (b) Plant Mask.
             (c) Stem Mask.
             (d) Stem Mask overlaid on original}
    \label{initial_process}
\end{figure}

\subsection{Overview}
Figure~\ref{Block_Diagram} shows a block diagram of our proposed method. Initially we use color correction, plant segmentation, and stem segmentation on the original RGB images. 
Color correction is used to address the image color inconsistency caused by the camera settings and acquisition conditions (\eg, lighting). 
Plant and stem segmentation are used to capture the plant shape information. 
These three initial image processing steps are required before estimating the wilting metrics. 
Figure~\ref{initial_process} shows an original plant image and the associated plant mask and stem mask.

All metrics can be categorized into color, shape (non-stem) based, and stem-based metrics.
Color and shape based metrics use color corrected images and the corresponding plant mask. 
Stem-based metrics require the plant mask and stem mask. 
To show the utility of our metrics, we also trained a random forest to predict a visual wilting score for a plant using our metrics.

\subsection{Initial Processing}

\begin{figure}[]
    \centering
    \subfloat[]{\label{Figure:input2}{\includegraphics[width = 0.35\textwidth]{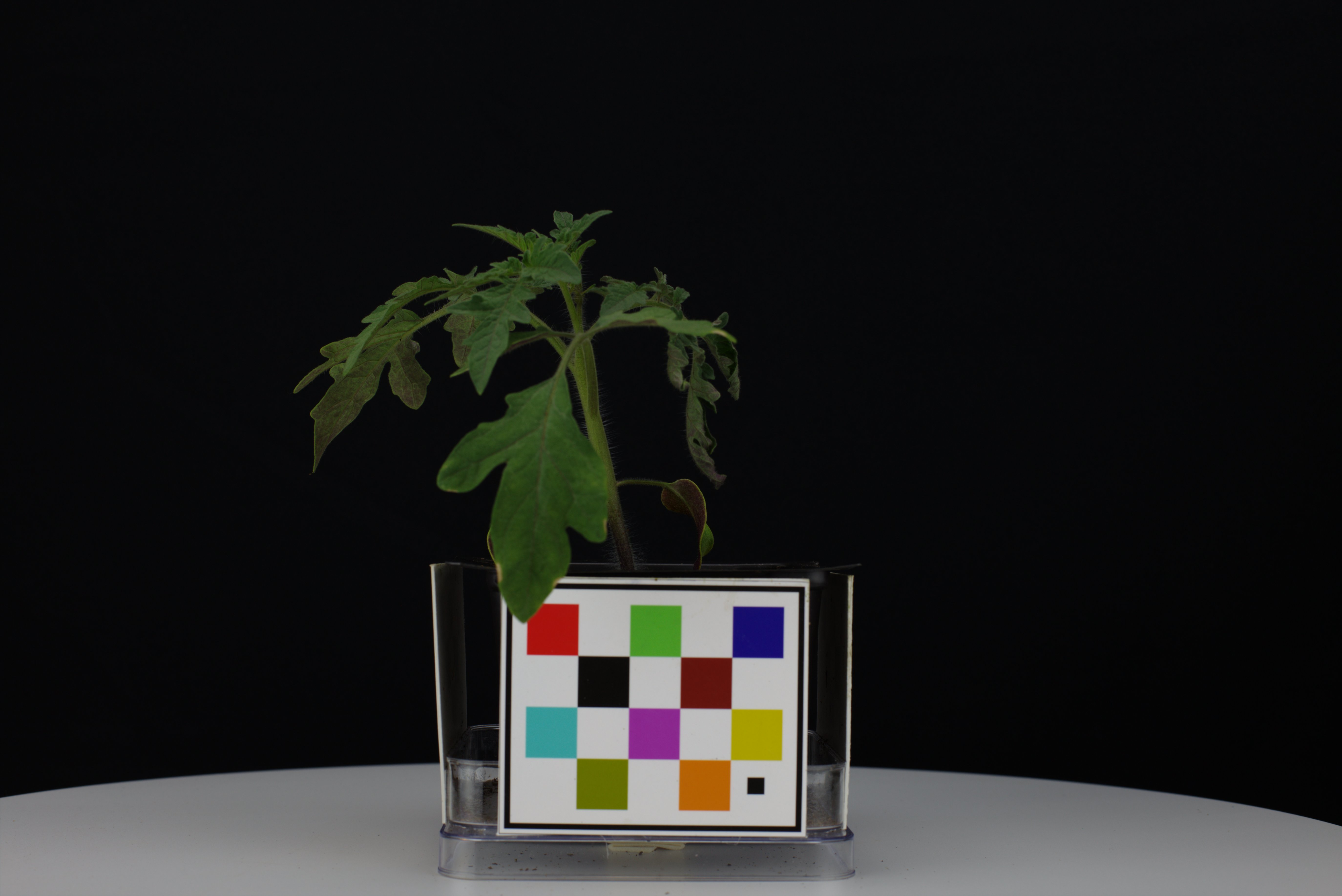}}}
    \qquad
    \subfloat[]{\label{Figure:prob2}{\includegraphics[width = 0.35\textwidth]{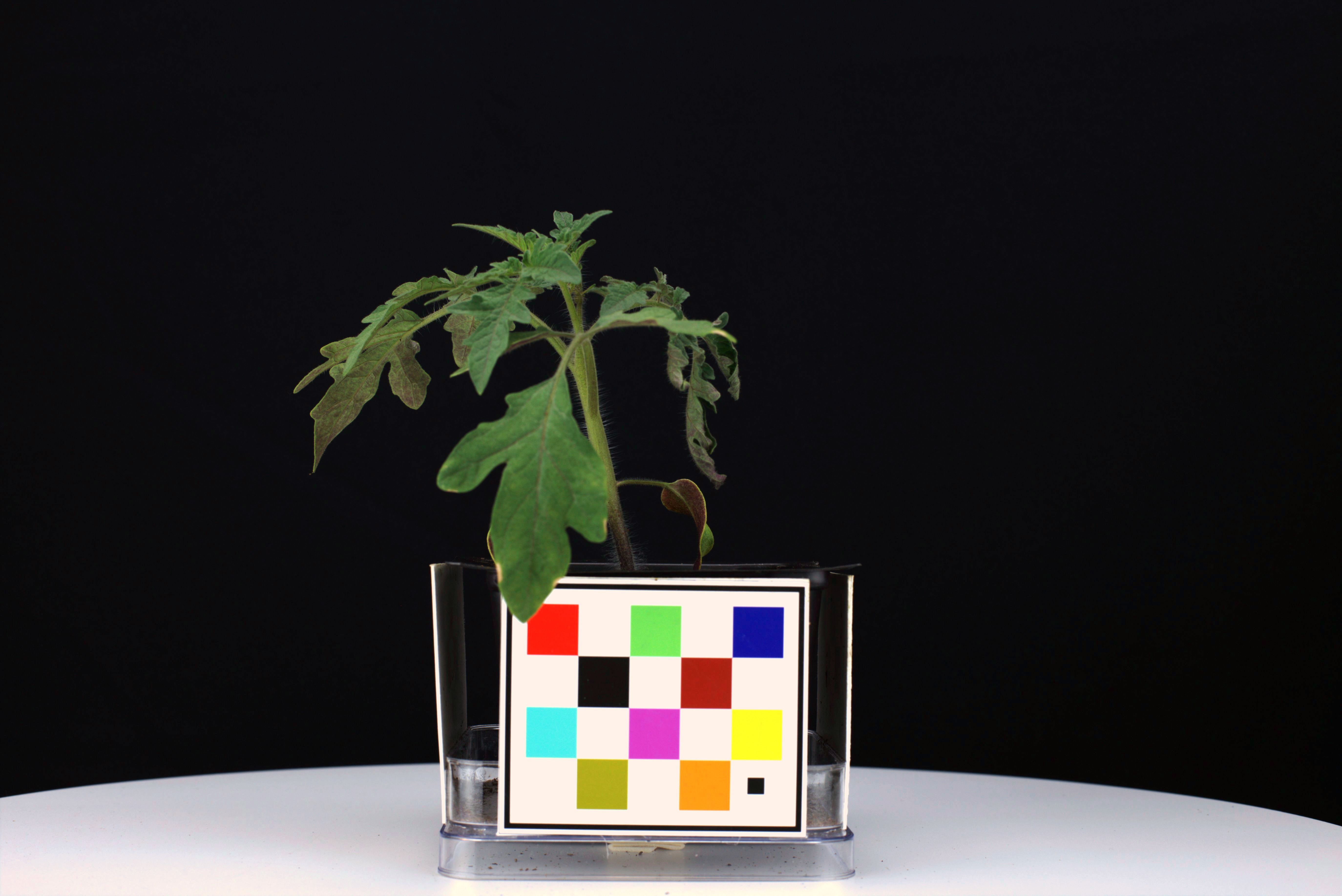}}}
    \qquad 

    \caption{Color correction.
             (a) Original Image
             (b) Color Corrected Image}

    \label{color_correct}
\end{figure}

\par \textbf{Color Correction:}
Since we are capturing wilting information across multiple days and for multiple plants, it is important to ensure that the lighting condition variations are suppressed as much as possible.
We color correct our images using a reference object that is known as a Fiducial Marker (FM) which can be seen in Figure \ref{color_correct}.
The FM is a colored checkerboard that has known physical dimensions and known colorimetric pixel values for each of its color squares.
We detect the FM in the image and estimate the average RGB pixel values for each of the color squares in the image.
Since we know the actual RGB values of the color squares, we can transform each pixel in the image to correct the color of each pixel.
Consider a $9\times3$ matrix $C_{\text{image}}$, which consists of the average R, G, and B pixel values for the $9$ color squares of the FM.
We also know the actual colorimetric  pixel values values of the color squares of the FM, 
represented by the $9\times3$ matrix $C_{\text{real}}$.
We estimate the $3\times3$ transformation matrix $T$:

\begin{align}
\label{eq:cc}
    C_{\text{real}} = C_{\text{image}} \times T \\
    \implies (C_{\text{image}}^TC_{\text{image}})^{-1}C_{\text{image}}^T \times C_{\text{real}} = T
\end{align}

An example of color correction is shown in Figure \ref{color_correct}.

\textbf{Plant Segmentation:}
After the color correction is completed, we segment the plant from the background.
After examining the images in multiple color spaces, we selected two channels that captured the plants in our image sufficiently: (1) the V channel from the HSV color space; and (2) the B* channel from the L*A*B* color space.
We obtain two preliminary plant segmentation masks from these channels by empirically determining separate thresholds for each image.
For our experiments, we used a threshold of $140$ for the V channel and $130$ for the B* channel, assuming the pixel values are between $0$ and $255$.
These preliminary binary masks are then combined with the logical `OR' function to obtain a single mask containing the information from both the V and B* channels.
To ensure any undesired objects in the mask are removed, we use sequential opening and closing operations~\cite{zhuang_1986} with a structuring element of ones in a $3 \times 3$ matrix to remove noise and fill holes in the mask.

\begin{figure}[!t]
    \centering
    \subfloat[]{\label{Figure:input2}{\includegraphics[width = 0.21\textwidth]{figures/r5-3dpi-wv18_ori.jpg}}}
    \quad
    \subfloat[]{\label{Figure:prob2}{\includegraphics[width = 0.21\textwidth]{figures/r5-3dpi-wv18_mask.jpg}}}
    \qquad \qquad \qquad    
    \subfloat[]{\label{Figure:input2}{\includegraphics[width = 0.21\textwidth]{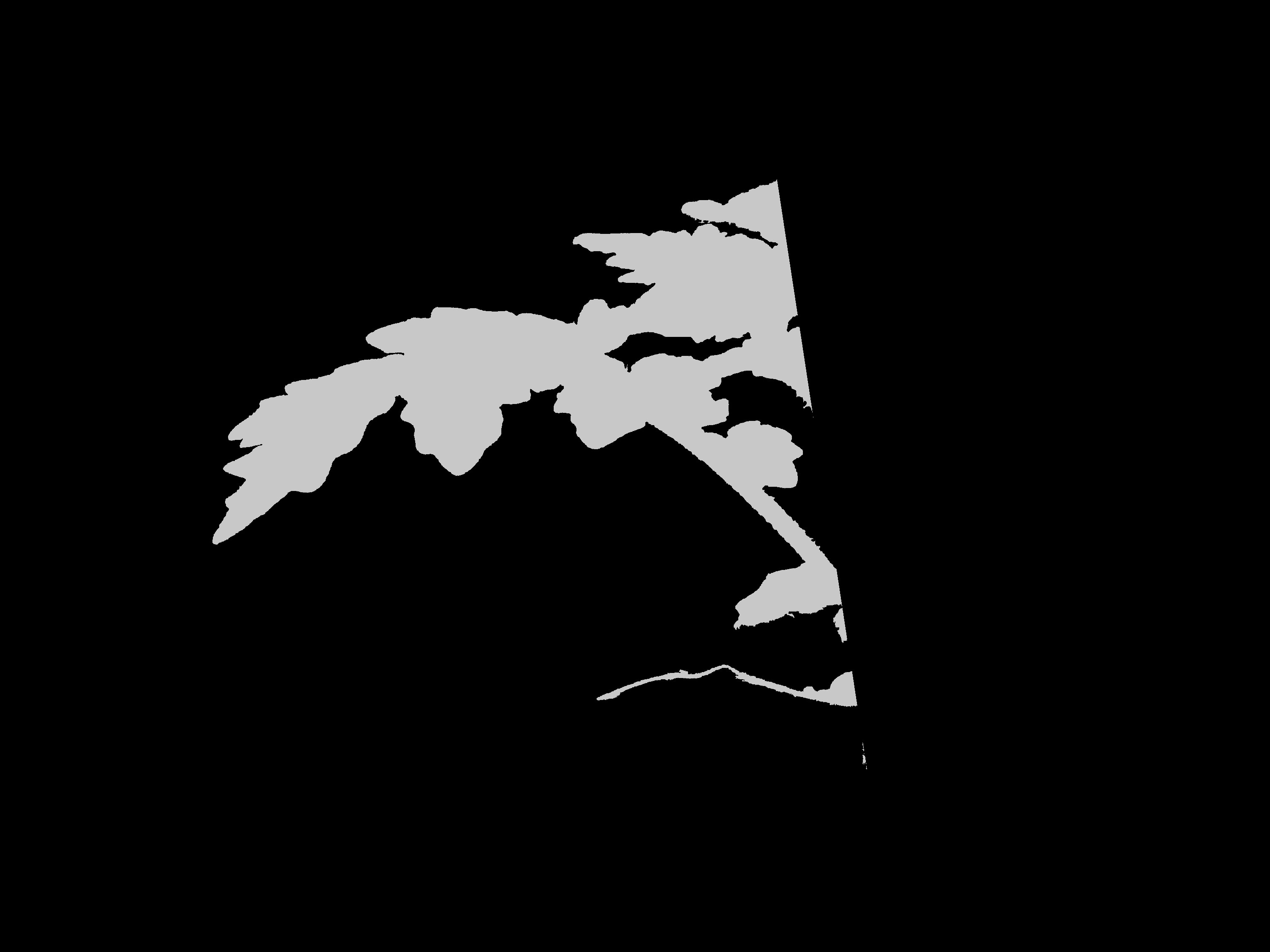}}}
    \quad
    \subfloat[]{\label{Figure:prob4}{\includegraphics[width = 0.21\textwidth]{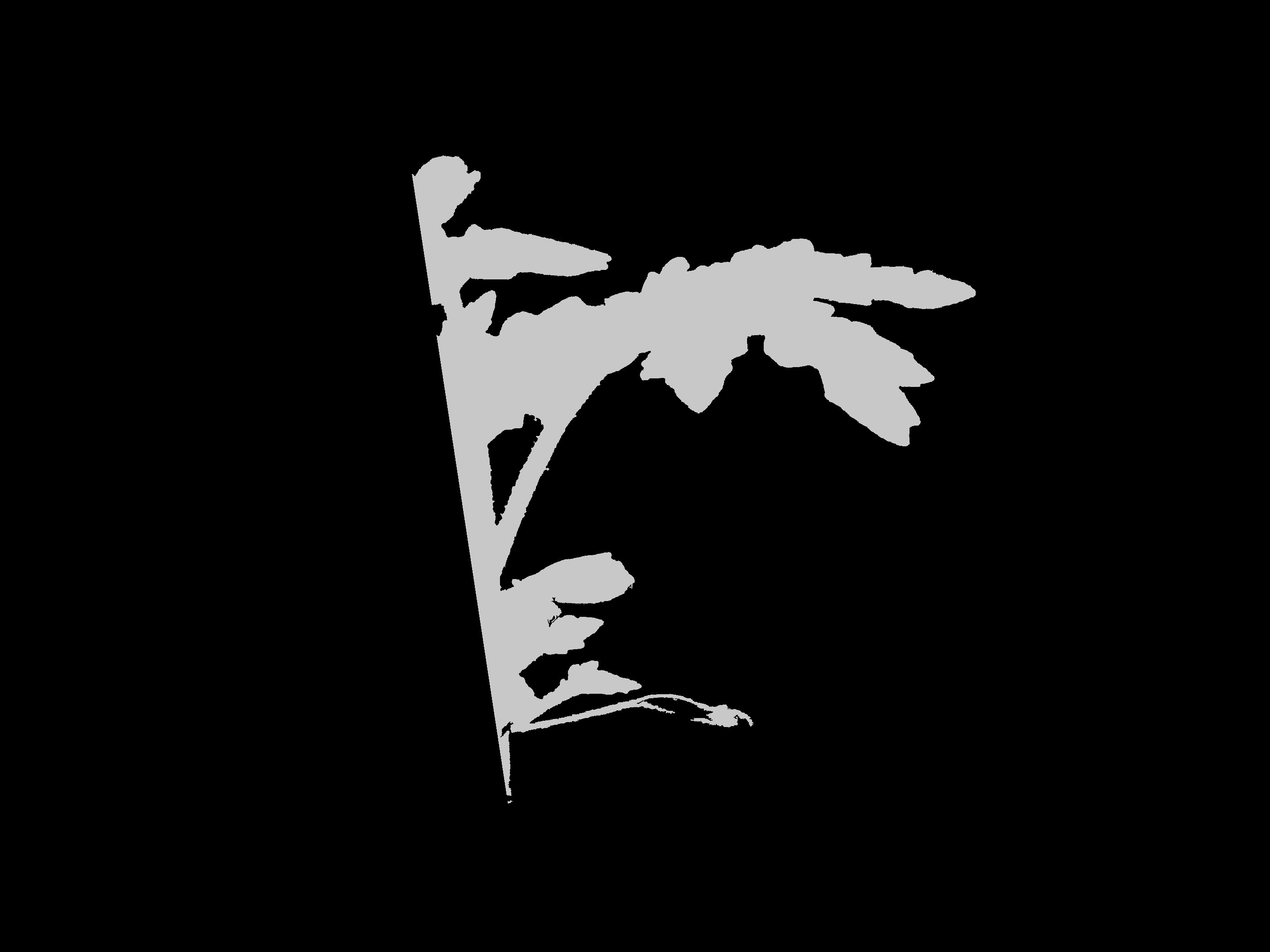}}}
    \qquad \qquad \qquad

    \subfloat[]{\label{Figure:input5}{\includegraphics[width = 0.21\textwidth]{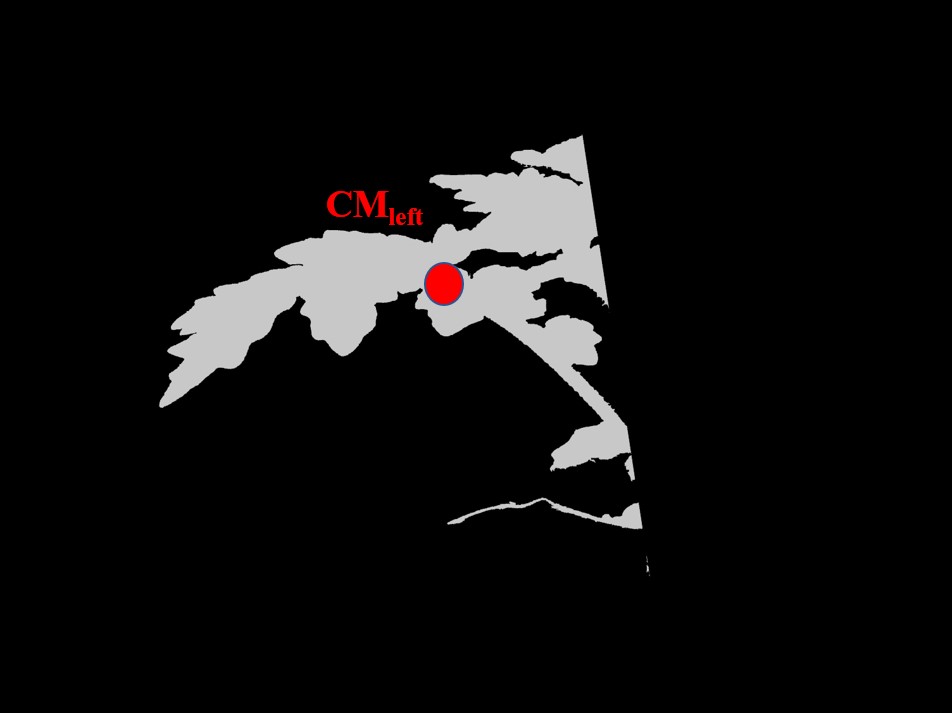}}}
    \quad
    \subfloat[]{\label{Figure:prob6}{\includegraphics[width = 0.21\textwidth]{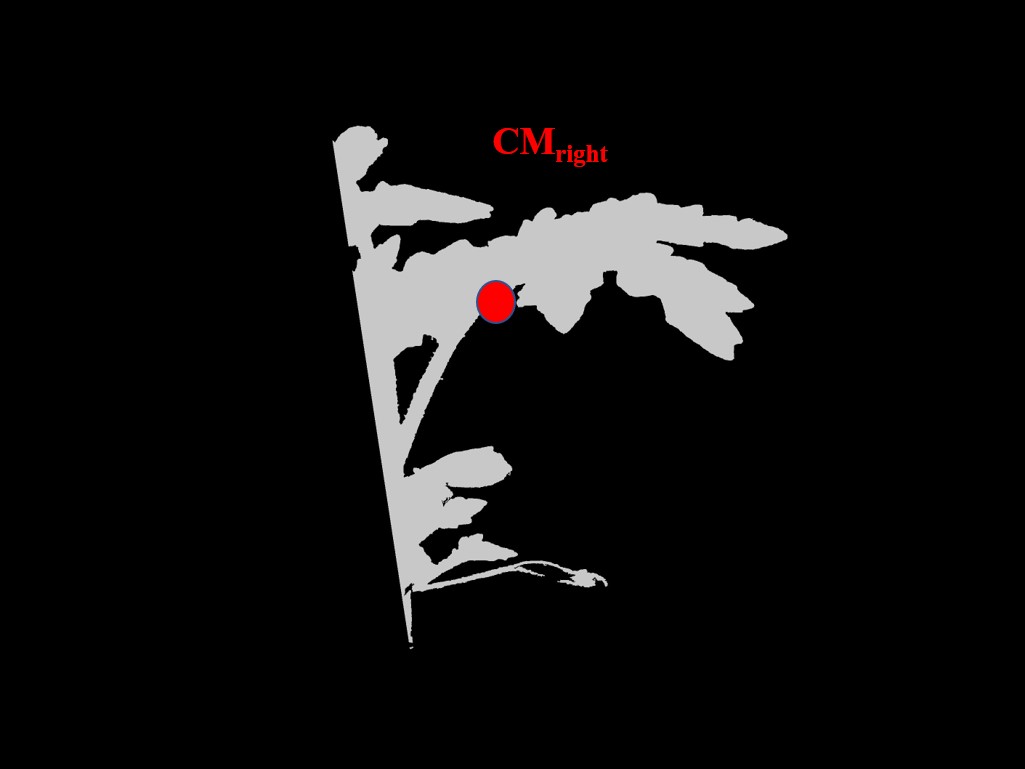}}}
    \qquad \qquad \qquad
    \caption{An example of the segmentation mask process: 
             (a) Original Plant
             (b) Plant Mask
             (c) Left plant Mask
             (d) Right Stem Mask
             (e) Left CM
             (f) Right CM
             }
    \label{mask_separation}
\end{figure}

\textbf{Stem Segmentation:}
For stem segmentation, we use a deep neural network-based solution similar to what was used by Yang \etal~\cite{yang2020}. 
A set of stem masks are manually labeled and used to train the stem segmentation networks. 
Two convolutional neural network (CNN)~\cite{vgg}  models, Mask R-CNN~\cite{maskrcnn} and U-Net~\cite{unet}, are used for our stem segmentation. 
The Mask R-CNN network is used for the majority of the stem segmentation tasks. 
For the plants where Mask R-CNN fails to detect stems, we use the U-Net to detect the stem. 
Mask R-CNN produces better quality~\cite{yang2020} masks compared to U-Net but it sometimes fails to detect the stem and outputs nothing. 
Since U-Net always produces a mask, it is used as a complementary stem segmentation network. 
To summarize, with the trained neural networks, our stem segmentation step takes the original image as input and generates a stem mask (Figure~\ref{initial_process}) as output.

\subsection{Color and Shape Based Metrics}
\textbf{Color:}
Using the plant segmentation masks described above, we next examine plant shape and color.
We are able to segment the relevant parts of the image, consisting of the plant, and examine the distribution of pixel values in the A* channel of the L*A*B* color space.
Since wilting plants tend to change color from green to brown, we want to capture the trend of the histogram  over time.
We compare the differences in the pixel distribution over time using the Bhattacharya distance~\cite{chattopadhyay2004}.
Our assumption is that resistant plants will have very small deviation from their original color distribution, while the wilting plants have a larger deviation.

\textbf{Shape:}
We capture the general shape of the plant using the convex hull~\cite{barber1996} around the plant mask.
We define two convex hull  metrics: perimeter of the plant object; and area of the convex hull of the plant object.
We can also find area of the plant object, width of the plant object, and height of the plant object using the plant mask $p_{msk}(x,y)$ (Figure~\ref{initial_process}).

\begin{figure}[]
	\centering
	\centerline{\includegraphics[width = 0.4\textwidth]{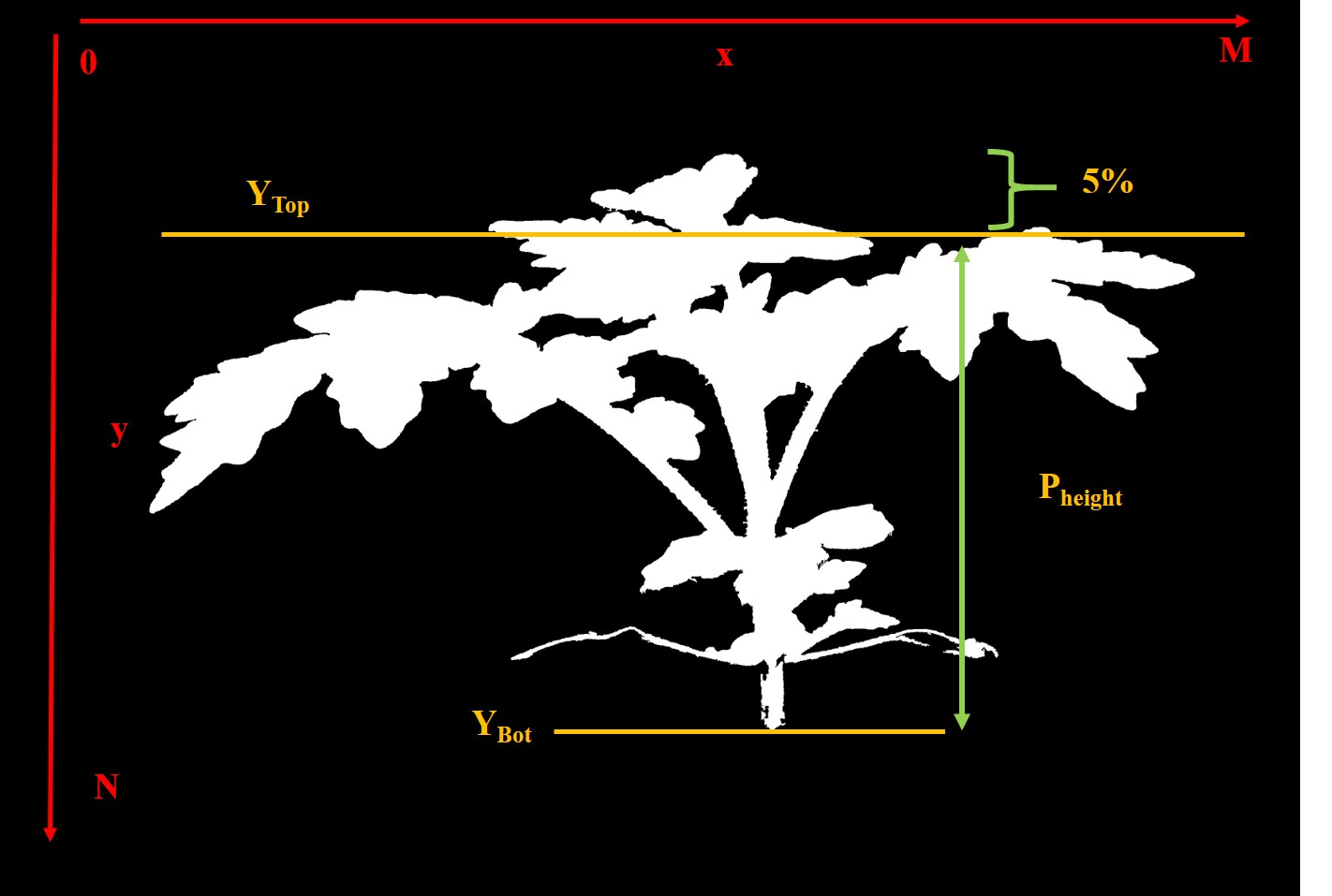}}
	\caption{Plant Height and Index Orientation}
\label{Plt_height}
\end{figure}

\textbf{Plant Height, Area, and Width:}
Our  indexing orientation is shown in Figure~\ref{Plt_height} and let the image pixel resolution be $C_{pres}$.
From the plant mask we can find plant area $P_{area}$ which is the total area of the plant material. 
\begin{equation}
	\label{plt_area}
    P_{area} = \sum_{x,y} p_{msk}(x,y)
\end{equation}
A horizontal profile $h_{hor}$ and a vertical profile $h_{ver}$ are estimated from $p_{msk}(x,y)$.
\begin{equation}
	\label{hor_prof}
    h_{hor}(y) = \sum_{x} p_{msk}(x,y)
\end{equation}
\begin{equation}
	\label{ver_prof}
    h_{ver}(x) = \sum_{y} p_{msk}(x,y)
\end{equation}
Plant width $P_{width}$ is defined as the difference between the leftmost pixel and rightmost pixel of the plant mask.

\begin{equation}
\begin{split}
P_{width} =& P_{max-r}-P_{max-l};
 \\where \quad & h_{ver}(P_{max-l}) \geq 1,\\ 
    & h_{ver}(P_{max-r}) \geq 1,\\
    &\sum_{i = 0}^{P_{max-l}-1} h_{ver}(i) = 0,\\
    &\sum_{i = P_{max-r}+1}^{N} h_{ver}(i) = 0\\
\end{split}
\end{equation}

\begin{figure}[]
	\centering
	\centerline{\includegraphics[width = 0.4\textwidth]{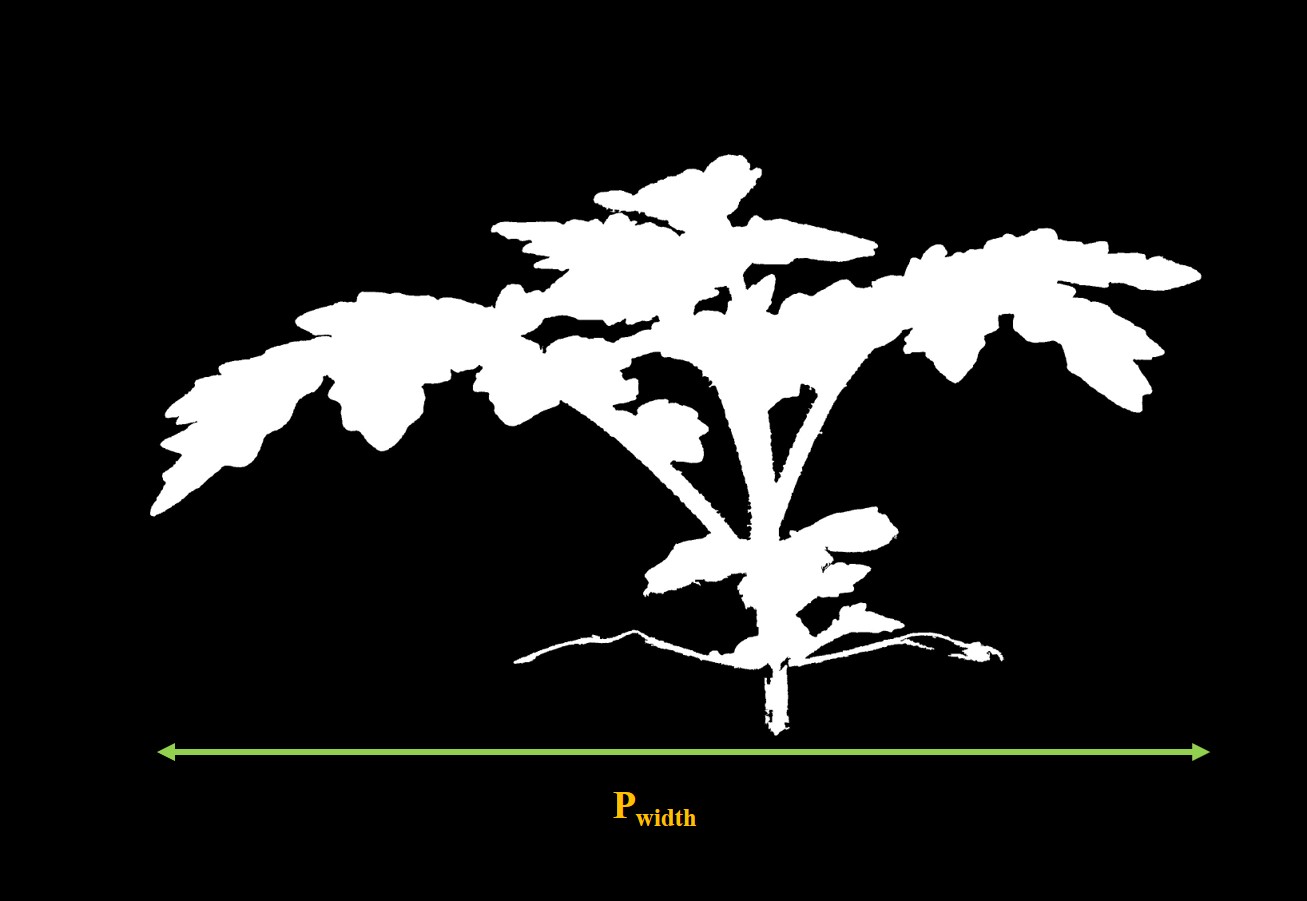}}
	\caption{Plant Width}
\label{Plt_Width}
\end{figure}

For plant height $P_{height}$, we remove the top $5\%$ of plant material and label the y-coordinate of the $5\%$ plant material cutoff line as $Y_{Top}$, where
\begin{equation}
	\label{plt_height}
    \sum_{i=0}^{Y_{Top}} h_{hor}(i) =  P_{area} \times 5\%
\end{equation}
We then define the upper edge of the pot as the bottom of the plant, denoting its average y-coordinate as $Y_{Bot}$. 
$P_{height}$ is defined as the difference between the $5\%$ plant material line and the bottom of the plant (Figure~\ref{Plt_height}). 
We remove the top $5\%$ of plant material so a small leaf at the top of the plant will not affect the plant height metric.
Figure \ref{Plt_height} and Figure \ref{Plt_Width} show a visualization of the $P_{height}$ and $P_{width}$. 
Plant Height, Area, and Width give an overview of the general shape of the plant.\par 

As a plant wilts, we expect almost every metric listed here to decrease.
The plant loses volume and turgidity, reducing its area, convex hull area, height, and width.
The convex hull perimeter may or may not decrease, depending on the final orientation of the plant in the image.

\subsection{Stem-Based Metrics}
Figure \ref{plt_evo} shows the change in plant material distribution for a healthier plant and a more infected plant. 
For a healthy plant in its early growth stage, the distribution of the plant material tends to be further away from the stem. 
Once the plants becomes infected and starts to wilt, the plant material will get closer to the stem due to the decreasing structural rigidity~\cite{Sun2017}. 
Based on our experimental observation and plant physics, we consider the stem of  plant to be a good reference point for plant material distribution estimation. 


\begin{figure}[]
	\centering
	\centerline{\includegraphics[width = 0.4\textwidth]{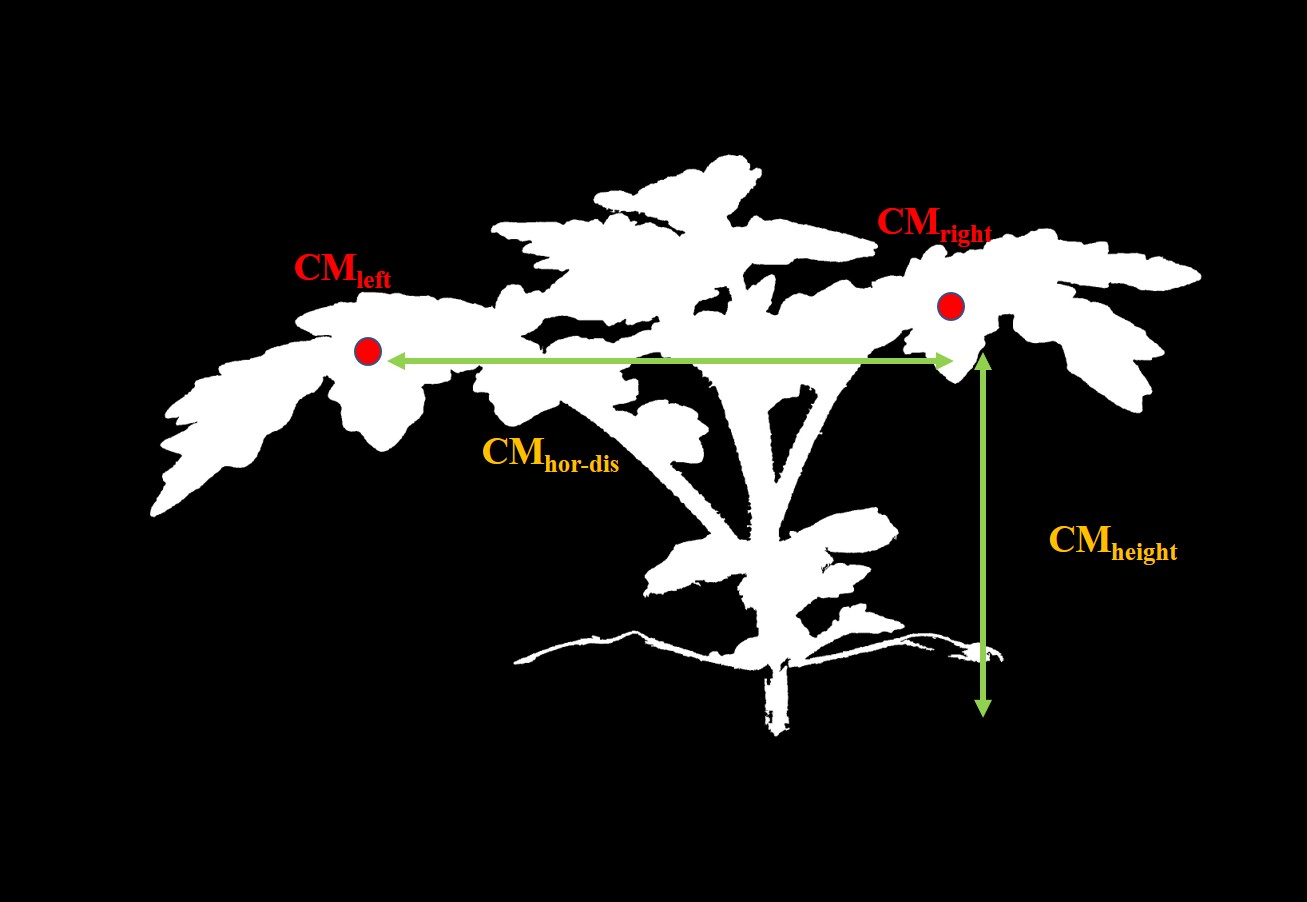}}
	\caption{Center of Mass (CM) Height and Distance}
\label{Plant_CM_h_and_d}
\end{figure}

The inputs required for stem-based metrics are the plant mask $p_{msk}(x,y)$, and stem mask $p_{stem}(x,y)$. 
Both the plant mask and stem mask are binary images with size $M \times N$ pixels. 

\textbf{Center of Mass:}
From the stem mask $p_{stem}$, we use linear regression~\cite{Steffensen2006} to form the function $s_{lin}(y)$. 
\begin{equation}
\begin{split}
s_{lin}(y) &= \alpha + \beta \cdot y \\
\alpha, \beta &= \arg \min_{\alpha, \beta}
\sum\limits_{\Bar{x},\Bar{y}} (\Bar{x} - \beta \cdot \Bar{y} -
\alpha)^{2}
\\ & \;\qquad\qquad\qquad \cdot p_{stem}(\Bar{x},\Bar{y}) \\
\end{split}
\end{equation}

Using $s_{lin}(y)$ we can then separate plant mask $p_{msk}(x,y)$ into a left plant mask $p_{l-msk}(x,y)$ and right plant mask $p_{r-msk}(x,y)$ (Figure \ref{mask_separation}). 
\begin{equation}
	\label{plt_height}
    p_{l-msk}(x,y)= 
    \begin{cases}
    p_{msk}(x,y) & \text{if $x \leq s_{lin}(y)$} \\
    0 & \text{else}
    \end{cases}
\end{equation}
\begin{equation}
	\label{plt_height}
    p_{r-msk}(x,y)= 
    \begin{cases}
    p_{msk}(x,y) & \text{if $x > s_{lin}(y)$} \\
    0 & \text{else}
    \end{cases}
\end{equation}
We then estimate the left center of mass ${CM}_{left}$ and right center of mass  ${CM}_{right}$. 
\begin{equation}
	\label{plt_height}
    {CM}_{left} = \left( \frac{\sum\limits_{x,y}^{}x\cdot p_{l-msk}(x,y)}{\sum\limits_{x,y}^{} p_{l-msk}(x,y)},\frac{\sum\limits_{x,y}^{}y\cdot p_{l-msk}(x,y)}{\sum\limits_{x,y}^{} p_{l-msk}(x,y)}
    \right)
\end{equation}
\begin{equation}
	\label{plt_height}
    {CM}_{right} = \left( \frac{\sum\limits_{x,y}^{}x\cdot p_{r-msk}(x,y)}{\sum\limits_{x,y}^{} p_{r-msk}(x,y)},\frac{\sum\limits_{x,y}^{}y\cdot p_{r-msk}(x,y)}{\sum\limits_{x,y}^{} p_{r-msk}(x,y)}
    \right)
\end{equation}

The x-coordinate difference between the left and right center of mass (CM) is set to be Center of Mass Horizontal Distance $CM_{hor-dis}$.
The average of the difference between the CM y-coordinates and the bottom of the plant $Y_{Bot}$ is defined as Center of Mass Height $CM_{height}$.
Figure \ref{Plant_CM_h_and_d} shows a visualization of the metrics. 

\begin{figure}[]
    \centering
    \subfloat[]{\label{Figure:input2}{\includegraphics[width = 0.21\textwidth]{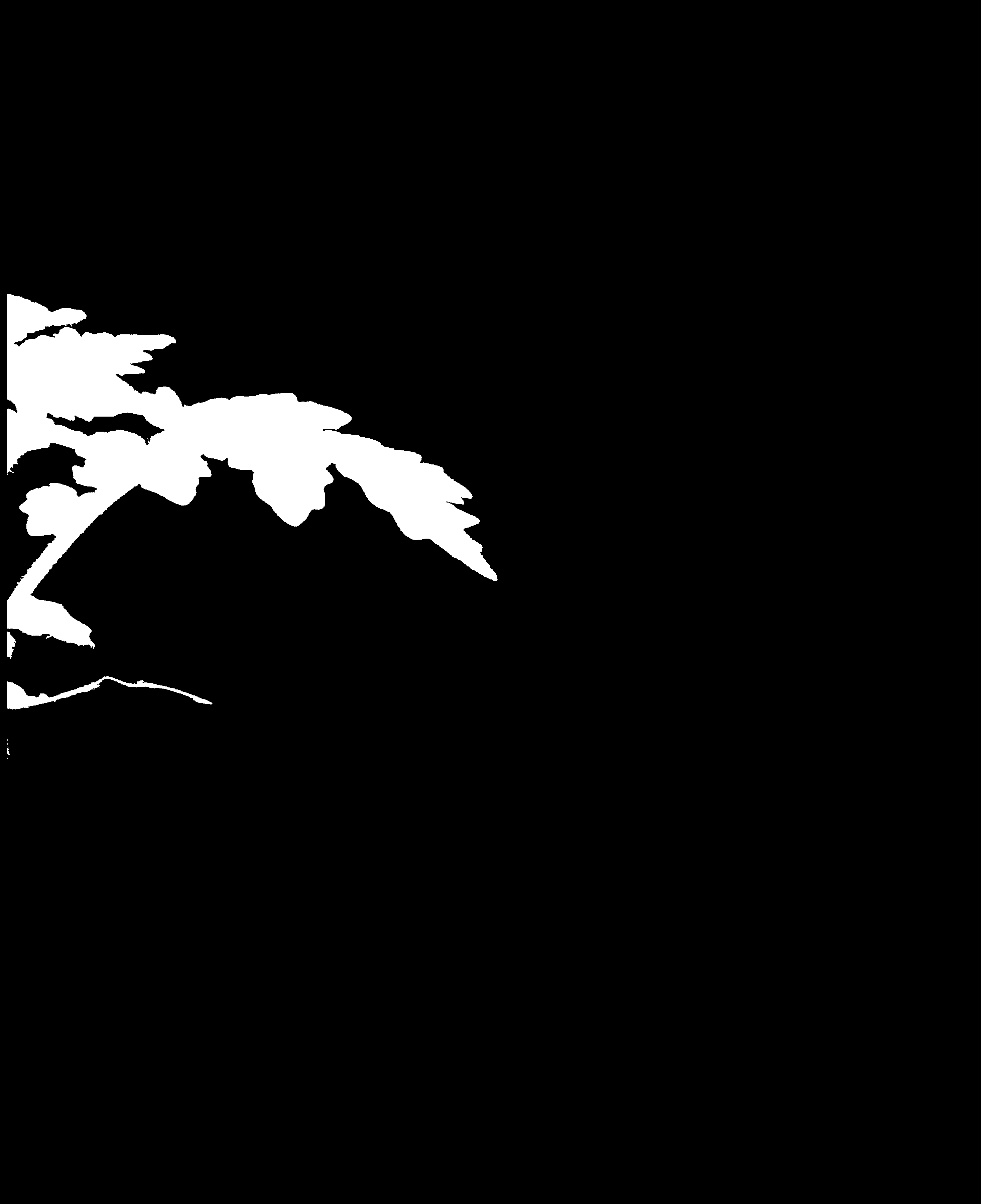}}}
    \quad
    \subfloat[]{\label{Figure:prob2}{\includegraphics[width = 0.21\textwidth]{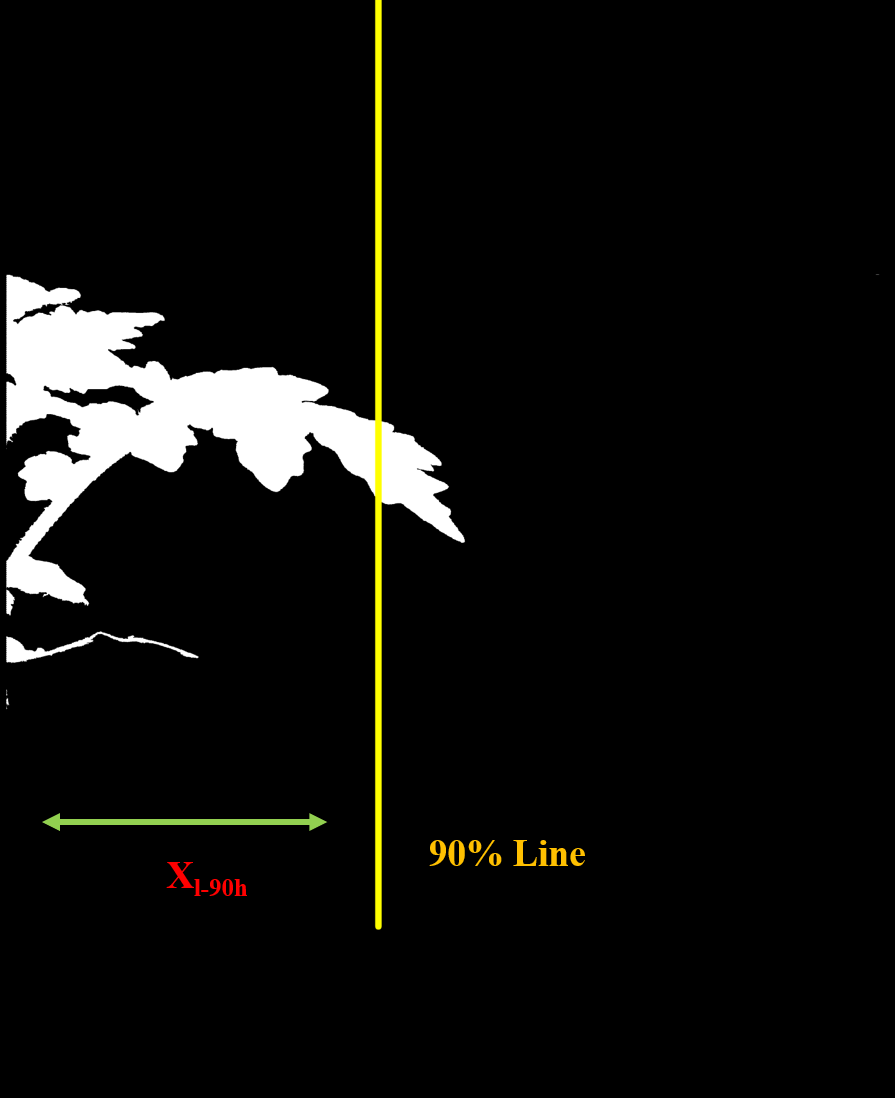}}}
    \qquad \qquad \qquad    

    \caption{
             (a) shifted and flipped left mask $R_{l-msk}(x,y)$ \\
             (b) Horizontal $90\%$ plant material line  $X_{l-90h}$}

    \label{hor_x_dis}
\end{figure}

\par\textbf{Vertical Distribution:}
The goal of the vertical distribution is to capture the plant material distribution along the y-axis for each half of the plant mask. 
We sample the distribution at $33\%$, $66\%$, and $90\%$. 
Lets  use $90\%$ as an example. 
For the left plant mask, we estimate the horizontal profile $h_{l-hprof}$. 
\begin{equation}
	\label{ver_prof}
    h_{l-hprof}(y) = \sum_{x} p_{l-msk}(x,y)
\end{equation}
Then we find the y-coordinate of the $90\%$ plant material line $Y_{l-90v}$, where
\begin{equation}
	\label{plt_height}
    \sum_{i=0}^{Y_{l-90v}} h_{l-hprof}(i) = \sum_{x,y} p_{msk}(x,y) \times 10\%
\end{equation}
The same steps are used to find $Y_{r-90v}$ using right plant masks. The average $90\%$ distribution $V_{90y}$ is defined as
\begin{equation}
	\label{plt_height}
    V_{90y} = \frac{Y_{r-90v} + Y_{l-90v}}{2}
\end{equation}

\textbf{Horizontal Distribution:}
The goal of the horizontal distribution is to capture the plant material distribution along the x-axis for each half of the plant mask. 
We sample the distribution at $33\%$, $66\%$, and $90\%$ as well.
Because the stem is not always vertical, we are not able to find a x-coordinate for the stem. 
We first do a horizontal shift to $p_{r-msk}(x,y)$ using the stem separation line, resulting in a shifted $R_{r-msk}(x,y)$. 
We then shift and flip $p_{l-msk}(x,y)$, resulting a flipped and shifted $R_{l-msk}(x,y)$ (Figure~\ref{hor_x_dis}).
\begin{equation}
	\label{plt_height} 
    R_{r-msk}(i,y) = p_{r-msk}(slin(y)+i,y)
\end{equation}
\begin{equation}
	\label{plt_height} 
    R_{l-msk}(i,y) = p_{l-msk}(slin(y)-i,y)
\end{equation}
The rest of the steps are the same as vertical distribution. 
Lets again use $90\%$ as an example. 
For left half plant masks, we estimate the horizontal profile $h_{l-vprof}$. 
Then we find the x-coordinate of the $90\%$ plant material line $X_{l-90h}$.  
Similar steps are taken to find $X_{r-90h}$ using right plant masks. 
The horizontal $90\%$ distribution $H_{90}$ is the sum of $X_{r-90h}$ and $X_{l-90h}$. 
Figure \ref{hor_x_dis} shows a example shifted mask$R_{l-msk}(x,y)$ and $X_{r-90h}$.
\begin{equation}
	\label{plt_height}
    H_{90x} = X_{r-90h} + X_{l-90h}
\end{equation}

All the shape-based and stem-based metrics above are then converted from pixels to metric units using the pixel resolution. 
Pixel resolution can be estimated from the dimensions of the Fiducial Marker.

\section{Experimental Design}
\label{exper}


We first demonstrate that our metrics can distinguish different plant genetics as well as how they been infected.  
We are also interested in comparing our methods to the way experts visually score a plant. 
Experts examine the plants visually eight days after inoculation
and rate each plant on its degree of wilting using  a value between 0 and 1.
We describe a random forest network method below to generate a predicted visual score from the proposed metrics. 
We then compare the predicted visual scores with visual scores assigned by expert plant scientists.

\textbf{Dataset:} Two species of tomato, Hawaii 7996 (HA) and West Virginia 700 (WV), are planted in a growth chamber with artificial lighting. 
HA is a tomato breed with strong resistance to \textit{Rs} and WV has weaker resistance compared to HA. 
Plants are separated into an experiment (Inoculated) group and control (Mock) group.  
The experiment (Inoculated) group was supplied with \textit{Rs}-infested water after germination. 
All plants are imaged at the day before inoculation, and three, four, five, and six days post inoculation (dpi). 

For all images, we use the same camera positioned at the same location and under controlled lighting conditions.
Fiducial markers are used for color correction. 
Each time a plant is imaged, eight side-view images are acquired from eight angles. 
For the inoculated group, expert observers assign a wilting score between 0 and 1 to each plant eight days after inoculation. 
In summary, there are 122 plants in the inoculated group. 
Of the 122 plants, 61 are HA and 61 are WV.
We have the expert visual scores for all 122 plants.
The mock group contains 36 plants (18 HA and 18 WV). 
Each image is $5496 \times 3670$ pixels with pixel resolution of $0.052$ cm / pixel.

\textbf{Statistical Analysis:} The goal of the statistical analysis is to show that our metrics can capture group differences. 
For example, in general, since HA should wilt less than WV after infection, we can check if our metrics are showing significant statistical difference between inoculated HA and WV. 
For each plant, we estimate our metrics from all eight side-images, using the average for each metric as the final value. 
There are cases where no metrics can be estimated from an image. 
This usually occurs if the plant is too small and plant segmentation or stem detection fails due to the image capture angle.
For these plants, the metrics estimated from the remaining sides are still averaged and used.

\begin{figure}[]
	\centering
	\centerline{\includegraphics[width = 0.5\textwidth]{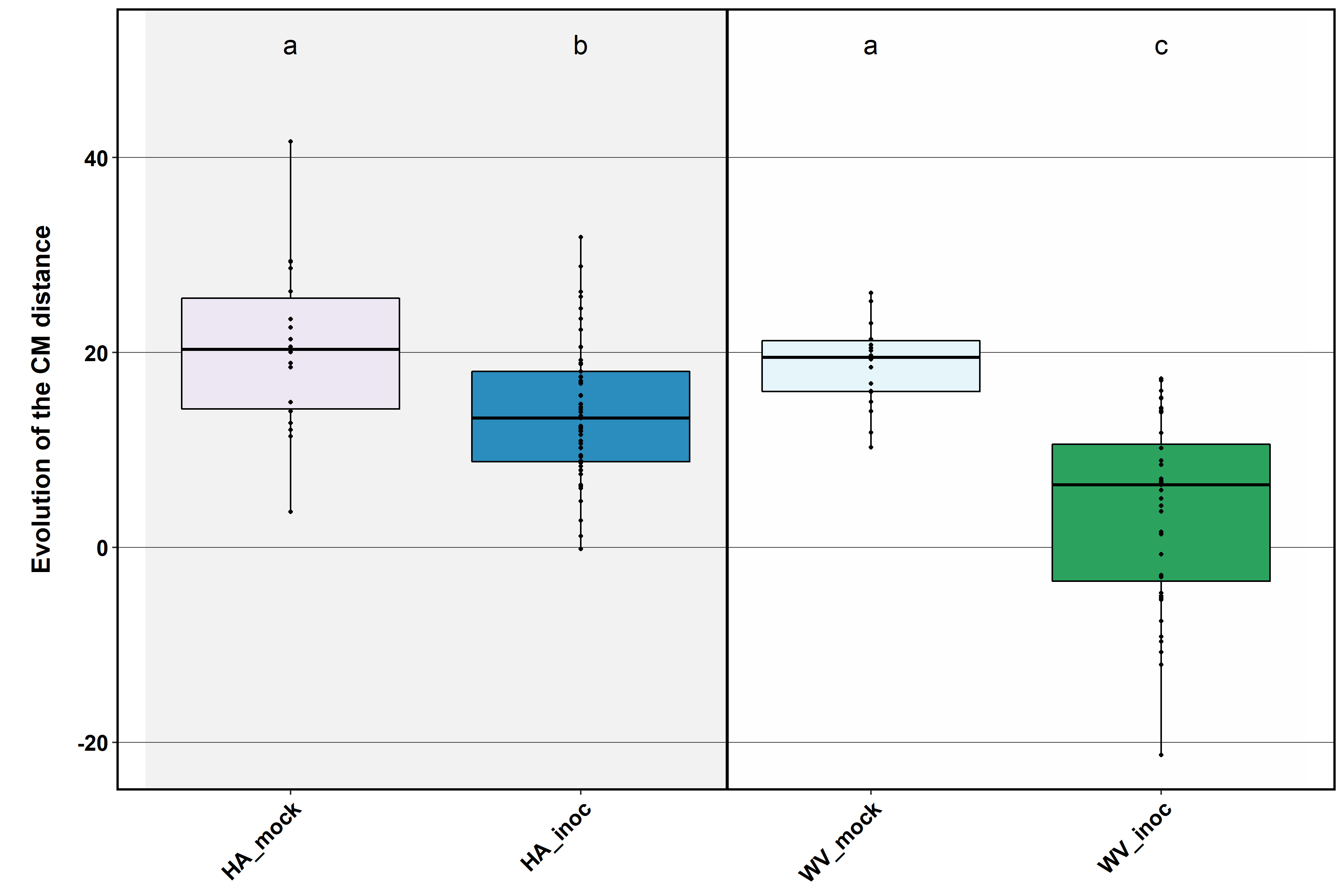}}
	\caption{Change of $CM_{hor-dis}$ from -1dpi to 3dpi in cm}
\label{h_CM_1-3_dif}
\end{figure}


\textbf{Random Forest:} We also compare our metrics to visual scores obtained by experts.
Our goal is to use our image-based metrics to generate a score that is similar to the expert's visual score.
Experts visually examine the plants eight days after inoculation, which is two days after the last images are captured.
They rate each plant on its degree of wilting using a score between 0 and 1, taking into consideration many features of the plant such as the turgidity of the petioles, the overall loss of plant mass, and the shift in color.

For our experiments, we treat the problem as a binary decision: either the plant has wilted or not.
Any plant with an expert wilting score above 0.5 is considered as ``wilted'', and any plant 
with an expert score below would be ``not wilted''.
We split the image-based metrics and the associated expert visual scores in a 6:4 ratio for training and testing.
We then train a random forest consisting of 1000 decision trees.
The random forest is evaluated by using the testing data to predict a visual score using our image-based metrics and then comparing it to the actual expert score.

\section{Discussion}

\subsection{Statistical Analysis}
\textbf{  $\mathbf{{CM}_{hor-dis}}$ Difference}:
In this section we compare the change of $CM_{hor-dis}$ from -1dpi to 3dpi. Figure \ref{h_CM_1-3_dif} shows the results for $CM_{hor-dis}$ difference between 3dpi and -1dpi.
Table \ref{CM_dptestresults} shows the Welch's t-test ~\cite{Tanton2005} results for some of the pairs.
From the statistical test results (Table \ref{CM_dptestresults}) and the distribution chart (Figure~\ref{h_CM_1-3_dif}), 
$\Delta CM_{hor-dis}$ shows no statistical difference between HA and WV when grown without \textit{Rs}. 
Once the plants are inoculated with bacteria, both HA and WV show decrease in $\Delta CM_{hor-dis}$.
This demonstrates that our metric is showing the effect of the inoculation on the plant as early as 3 dpi. 
 
We can also observe that once the plants are inoculated, the more \textit{Rs}-resistant HA plants have a much smaller decrease in $\Delta CM_{hor-dis}$ than WV plants.
This is a indicator that our metric is capturing the plant resistance to \textit{Rs} rather than some other inherent genetic difference because $\Delta CM_{hor-dis}$ only shows a statistical difference between HA and WV when the plants are inoculated with \textit{Rs}. 
Overall, difference of $CM_{hor-dis}$ matches our prior knowledge about the resistance of the plants, and it demonstrates the effect of inoculation at an early stage.

\begin{figure}[]
	\centering
	\centerline{\includegraphics[width = 0.5\textwidth]{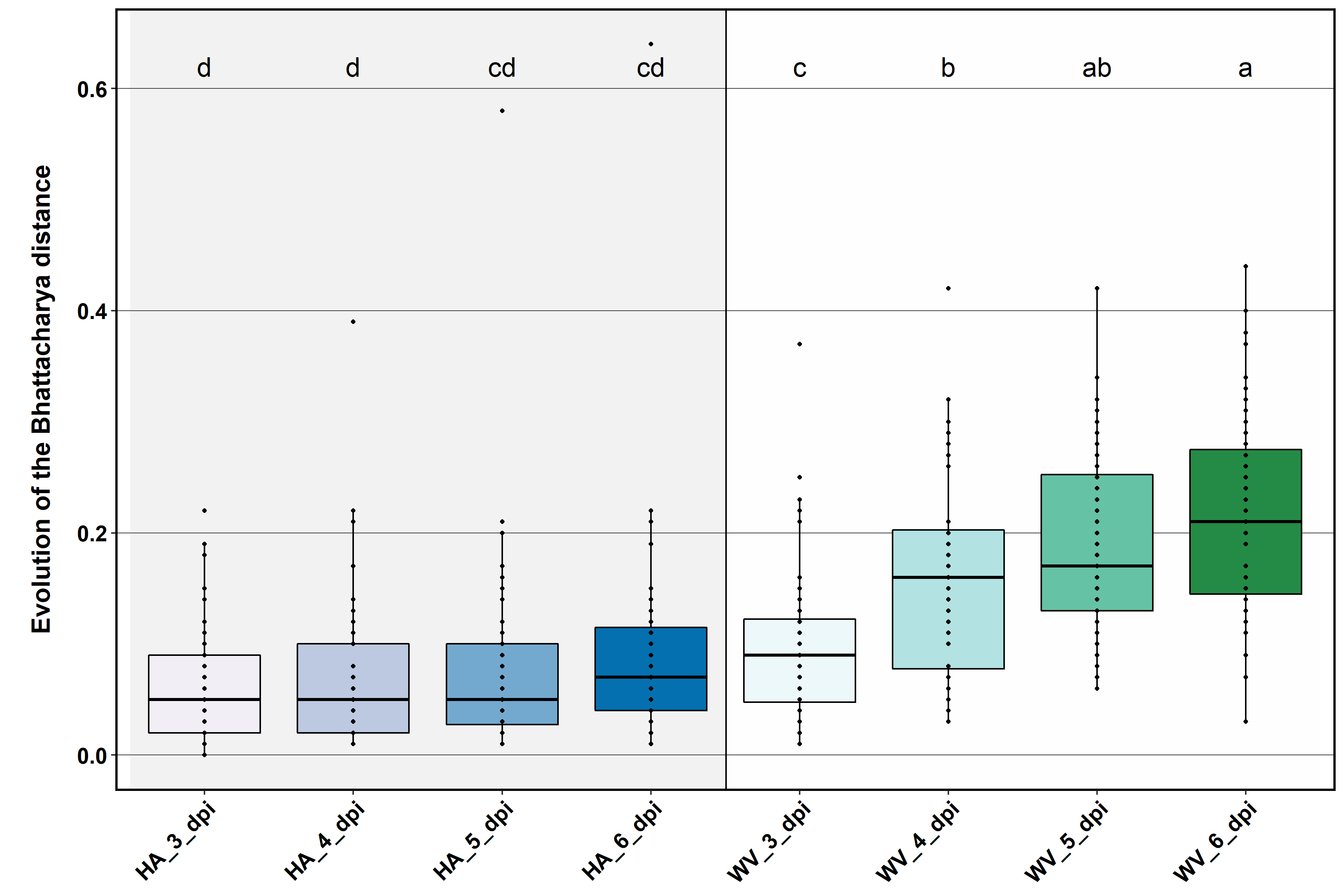}}
	\caption{Bhattacharya Distance for inoculated HA and WV plants}
\label{BD_I_ALL}
\end{figure}
\textbf{Bhattacharya Distance}:
In this section we examine the impact of bacterial wilt in the inoculated group.
Figure \ref{BD_I_ALL} shows the Bhattacharya Distance (BD)~\cite{chattopadhyay2004} of pixel color distributions on different dpis for inoculated HA and WV plants. 
Here, the BD measures the difference in the pixel distribution of color for each day post inoculation from the pixel distribution of color pre-inoculation. 
The Kruskal-Wallis~\cite{Tanton2005} test for inoculated HA plants returns a $p$-value of $0.235$ and for inoculated WV plants, it returns a $p$-value of $3.52e-12$. 

As we can see from Figure \ref{BD_I_ALL}, the distribution of the color pixels in inoculated HA plants do not have significant changes and the distribution of the color pixels in inoculated WV plants continues to deviate further from the distribution pre-inoculation. 
The results show that BD can be a good indicator of bacterial wilt since the more \textit{Rs}-resistant plants (HA) have no significant pixel distribution shift while the WV plants do.



\begin{table}[h]
\centering
\begin{tabular}{lll}
\toprule
\textbf{Tests} &  \textbf{p-value} \\\midrule 
Inoc HA vs. Inoc WV & $\mathbf{1.57e-7}$   \\
Inoc HA\ vs. Mock WV & $\mathbf{0.0067}$   \\
Inoc HA vs. Mock HA & $\mathbf{0.0078}$   \\
Mock HA vs. Mock WV & $0.45$   \\
\bottomrule
\end{tabular}
\caption{Center of Mass (CM) Distance $t$-test results}
\label{CM_dptestresults}
\end{table}

\subsection{Random Forest}

Recall that we have obtained expert visual scores representing the degree of wilting for each of the plants eight days post inoculation.
Our goal is to use our image-based metrics to generate a score that is similar to the expert visual score. 
We want to demonstrate that our metrics contain enough information to reach the same conclusions as a trained expert. 
As discussed previously, we train a random forest consisting of $1000$ decision trees to generate a score using our image-based metrics.
We evaluate the scores output by the random forest using Precision, Recall, and F1 Score~\cite{powers_2011}:

\begin{equation}
	\label{precision}
    \text{Precision} = \frac{\text{TP}}{\text{TP}+\text{FP}},
\end{equation}

\begin{equation}
	\label{recall}
    \text{Recall} = \frac{\text{TP}}{\text{TP}+\text{FN}}.
\end{equation}

\begin{equation}
	\label{f1}
    \text{F1 Score} = \frac{2 \times \text{Precision}\times \text{Recall}}{\text{Precision}+\text{Recall}}
\end{equation}

These metrics can be used for either class of ``wilted'' or ``not wilted'', where the definitions of True Positive (TP), False Positive (FP), and False Negative (FN) will change depending on the class.
The results are shown in Table \ref{rf_acc}.
For the testing dataset of 48 plants, the random forest model was able to use our metrics to predict the final state of the plant with an accuracy (F1 Score) of $100\%$.
The predicted visual scores from our metrics match the scores provided by the experts for every plant.
Note that the expert visual scores are based on the state of the plant at eight days post inoculation.
The performance of our model is achieved using metrics from images up to six days post inoculation, meaning that the random forest is able to predict the final state of the plant two days in advance.

\begin{table}[h]
\centering
\begin{tabular}{llll}
\toprule
Class & Precision & Recall & F1\\
\midrule 
Wilted & 1.00 & 1.00 & 1.00\\
Not Wilted & 1.00 & 1.00 & 1.00 \\
\bottomrule
\end{tabular}
\caption{Random Forest Classification Results}
\label{rf_acc}
\end{table}

\section{Conclusion and Future Work}
In this paper, we proposed image-based metrics for estimating the amount of wilting in a diseased plant. 
We demonstrate that our metrics are able to distinguish between diseased plants and healthy plants of the same species. 
We can also distinguish between resistant plant species and less resistant plant species. 
We also use our metrics to predict visual wilting scores using a random forest. 
We conclude that our metrics are effective for measuring bacterial wilt. 
For future work we will extend our methods to other species and investigate new features to improve our wilting estimation.

{\small
\bibliographystyle{ieee_fullname}
\bibliography{ref}
}

\end{document}